\documentclass[letterpaper, 10 pt, conference]{ieeeconf}  % Comment this line out
\IEEEoverridecommandlockouts                              % This command is only
\usepackage{graphics} % for pdf, bitmapped graphics files
\usepackage{epsfig} % for postscript graphics files
\usepackage{mathptmx} % assumes new font selection scheme installed
\usepackage{times} % assumes new font selection scheme installed
\usepackage{amsmath} % assumes amsmath package installed
\usepackage{amssymb}  % assumes amsmath package installed
\usepackage{graphicx}
\usepackage{tabto}
\usepackage{hyperref}
\usepackage{amsmath}
\usepackage{makecell}
\usepackage{algorithmic}
\usepackage[linesnumbered,ruled]{algorithm2e}
\usepackage{cite}
\title{\LARGE \bf
Fuzzy Logic Control for Indoor Navigation of Mobile Robots
}

\author{Akshay Kumar$^{*}$, Ashwin Sahasrabudhe$^{*}$ and Sanjuksha Nirgude$^{*}$% <-this % stops a space
\thanks{$^{*}$The authors are affiliated with the Department of Robotics Engineering, Worcester Polytechnic Institute (WPI), MA, USA}%
}

\begin{document}

\maketitle
\thispagestyle{empty}
\pagestyle{empty}

%%%%%%%%%%%%%%%%%%%%%%%%%%%%%%%%%%%%%%%%%%%%%%%%%%%%%%%%%%%%%%%%%%%%%%%%%%%%%%%%
\begin{abstract}

Autonomous mobile robots have many applications in indoor unstructured environment, wherein optimal movement of the robot is needed. The robot therefore needs to navigate in unknown and dynamic environments. This paper presents an implementation of fuzzy logic controller for navigation of mobile robot in an unknown dynamically cluttered environment. Fuzzy logic controller is used here as it is capable of making inferences even under uncertainties. It helps in rule generation and decision making process in order to reach the goal position under various situations. Sensor readings from the robot and the desired direction of motion are inputs to the fuzzy logic controllers and the acceleration of the respective wheels are the output of the controller. Hence, the mobile robot avoids obstacles and reaches the goal position. 

\end{abstract}


%%%%%%%%%%%%%%%%%%%%%%%%%%%%%%%%%%%%%%%%%%%%%%%%%%%%%%%%%%%%%%%%%%%%%%%%%%%%%%%%
\section{INTRODUCTION}

Autonomous navigation systems have distinct approaches to trajectory generation, path planning, control and required computation to execute the tasks for self-driving vehicles and mobile robotic platforms. Unlike self-driving vehicles, autonomous mobile robots for indoor as well as outdoor applications do not have a specific road-like path to maintain while moving ahead, thereby having the independence to plan and track any feasible and easy path to the target location while avoiding obstacles and satisfying other dynamic constraints.

Over the years, several control techniques have been deployed for efficient performance of these mobile robotic platforms. These techniques range from classical methods like PID control, trajectory control and position control to sophisticated methods like Model Predictive Control and Fuzzy Logic Controller \cite{c1}, \cite{c2}. PID Control technique is the easiest of all, but suffers from issues of tuning and robustness; the major deterrent to its use in any real-time high fidelity demanding problem, like mobile robot platforms. Other techniques like trajectory and position control work in environments without disturbances and/or unprecedented possibilities. However, Receding Horizon-Model Predictive Control shows promising results but it is mathematically quite expensive, making the implementation tough.  

To accommodate such limitations, we used the Fuzzy Logic Controller \cite{c3}, \cite{c4} for navigation of a mobile robot platform of TurtleBot2 in Gazebo simulation environment. A fuzzy control system runs on fuzzy logic (no hard decisions) by considering analog inputs as continuous logical variables ranging between 0 and 1 instead of strictly 0  or strictly 1. It essentially means asserting conditions to be "partially true/false" instead of "true/false".  

% \subsection{Motivation}

% Mobile robot navigation in indoor as well as outdoor scenarios has been a topic of intricate research over the years. Given the discord in transfer of the latest state-of-the-art technologies for self-driving vehicles to these mobile robots, we focus our project to develop and refine techniques specifically applicable to the latter systems. Though the use of fuzzy logic control has been long in use, the methods of tuning and further exploration into refining the technique is still sought after. We propose this topic to explore how optimization techniques combined with fuzzy control to enhance performance in static as well as dynamic environments.      

\subsection{Literature Survey}
Fuzzy Logic controller has been used many times for control of mobile robots. In \cite{c6} both the navigation and obstacle avoidance approaches are used. The method is applied on a non-holonomic mobile robot. In the paper \cite{c7} the authors have used fuzzy logic controllers with various types of inputs like sonar, camera and stored map. An application of fuzzy logic controller is proposed for indoor navigation in the paper \cite{c8}. In this paper wheeled mobile robots(WMR) are used. Another application of the fuzzy logic controller for indoor navigation is presented in the paper \cite{c9}.In this paper visual sensors are used to guide the robot to the target, but they do not use FLC for obstacle avoidance. 

\section{FUZZY LOGIC}

The section is divided into two subsections - subsection A discusses the general approach to fuzzy control while subsection B explains the exact techniques for inference and processing used to implement the proposed controller. 

\subsection{Fuzzy Control Approach}
\begin{figure}[ht]
  \centering
  \includegraphics[width = 8.5cm]{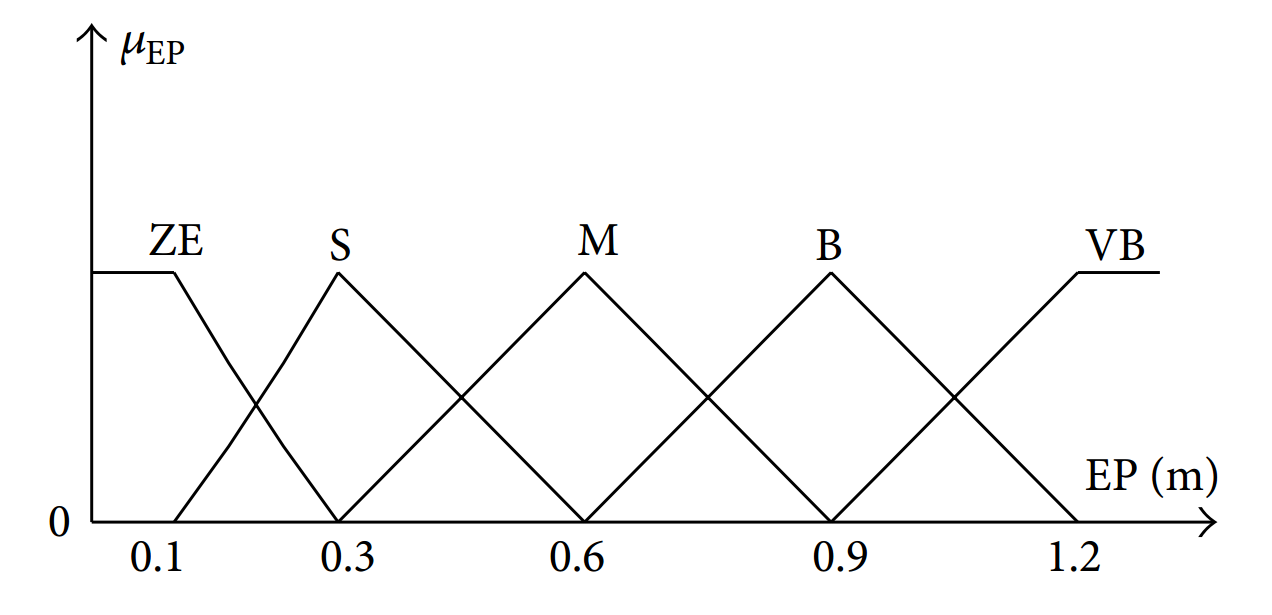}
   \caption{Example Membership Function Representation}
   \label{fig2}
\end{figure}

Fuzzy logic theory is a solution to control mobile robots. The basic structure of a fuzzy logic controller is composed of three steps. The first step is fuzzification which transforms real values inputs and outputs  into grade membership functions for fuzzy control terms. An example membership function generation setup is shown in Figure \ref{fig2}. The second step is the inference which combines the facts acquired from the fuzzification step and conducts a reasoning process. The basic fuzzy rules depend on the information acquired which is then reasoned using the 'If-antecedents-then-conclusion' rule. The last step is the defuzzification which transforms the subsets of the outputs which are calculated by the inference step. 

We use a combination of two fuzzy logic controllers to complete our task. For navigation a Tracking Fuzzy Logic Controller (TFLC) would be used and an Obstacle Avoiding Fuzzy Logic Controller(OAFLC) would be used for avoiding unknown obstacles in the cluttered environment. The lack of information of the environment makes it a challenging problem to navigate. The TFLC and OAFLC are combined to navigate the robot to the target along a collision free path. The algorithm starts with TFLC and whenever there is an obstacle in the path, it switches to OAFLC. The output of this algorithm are velocities of left and right wheels.

TFLC helps to move the robot to the target smoothly by taking the distance and the angle between the robot and the target as its inputs. OAFLC is used to generate a control signal in order to avoid obstacles. The inputs to the OAFLC are the distances from the obstacles at certain angles from the robot. These distances are acquired from the  depth sensor of Kinect Sensor on TurtleBot. The velocities of the left and right wheels are calculated using the defuzzification step.

\subsection{Fuzzy Techniques}

Here, we discuss the techniques used for the two important implementations of the Fuzzy Logic Controller - the Fuzzy Inference system and the defuzzification technique. The Takagi-Sugeno-Kang fuzzy inference technique and the Centroid defuzzification methods are used to implement our proposed controller. The TSK approach computes the output of the If-Else rules as a linear expression made up of weighted conditional components. Elaborately, the FIS setup processes all If-Else conditional statements with the weights generated on the basis of the membership functions and computes a new weight for execution of the condition. Further, the Centroid defuzzification process computes a normalized weight distribution for conditions and thereafter their weighted sum to generate final numerical output values. These techniques have similar implementation for the Tracking FLC as well as the Obstacle Avoidance FLC. 

 \section{METHODOLOGY}

In order to implement Fuzzy Logic Controller on a mobile robot platform, the TurtleBot2 robot platform is being used. Gazebo simulator with ROS support is being used for simulation, testing and environment creation platform. 
This methodology section has been further divided into subsections that explain the hardware setup, the software design, environment setup, fuzzification of sensor data, controller implementation methodology and final implementation nuances of the proposed system.

\subsection{TurtleBot2 Hardware}

\begin{figure}[ht]
  \centering
  \includegraphics[width = 8.5cm]{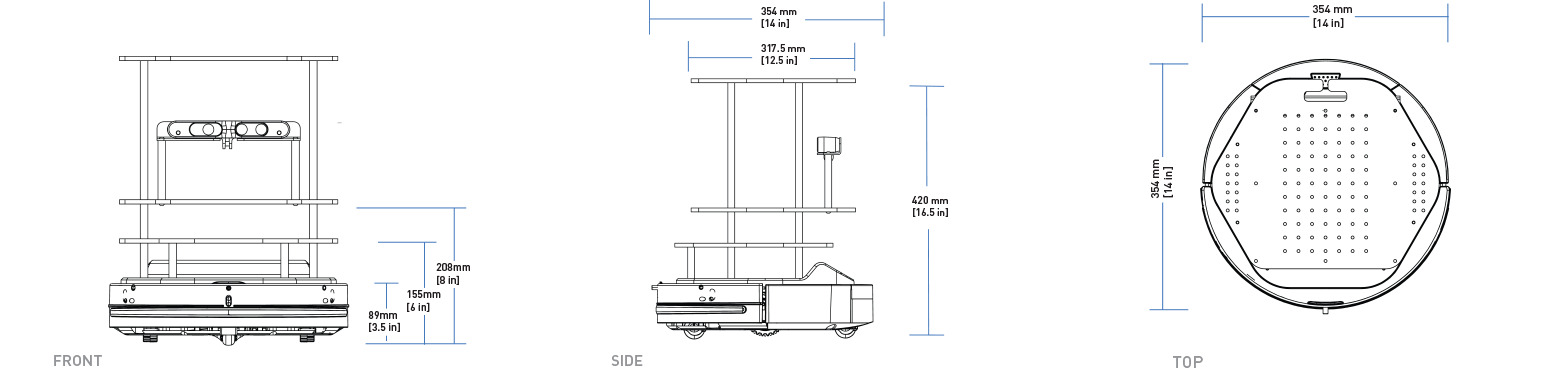}
   \caption{Specifications of TurtleBot2}
   \label{fig1}
\end{figure}

Figure \ref{fig1} shows the CAD specifications of the TurtleBot2 mechanical model and design.

The TurtleBot2 is an extended work placed atop a standard differential drive mobile base from Kobuki. It has several sensors like the bump sensor and cliff sensors on the base. The IMU sensor on the base observes the angular heading and senses the variations over motion. Table 1 mentions the several hardware specifications of the Kobuki base being used. 

\begin{table}[ht]
\caption{Hardware Specifications of the Kobuki Base}
\label{table_example}
\begin{center}
\begin{tabular}{|l|l|}
\hline
Max. Linear Velocity & 70 cm/s \\
\hline
Max. Rotational Velocity & 180 $\theta/s$ (>110 $\theta/$s gyro performs poorly) \\
\hline
Payload & 5 kg (hard floor), 4 kg (carpet) \\
\hline
Threshold Climbing & Climbs thresholds of 12 mm or lower \\
\hline
Rug Climbing & Climbs rugs of 12 mm or lower \\
\hline
Expected Operating Time & 3/7 hours (small/large battery) \\
\hline
Expected Charging Time & 1.5/2.6 hours (small/large battery) \\
\hline
\end{tabular}
\end{center}
\end{table}

The TurtleBot2 version used here has a Asus Xion Pro Live mounted for perception. We use the depth sensing and consequent conversion of the same into a 2D laser scan to learn about the presence of obstacles for navigation. It has $58.5^o$ and $48.0^o$ horizontal and vertical angular ranges of view respectively. Its linear range of view is 80cm to 4m in far mode and 40cm to 3m in near mode. Given the large area of observation, we are able to create several levels of fuzzy logic for control.

\subsection{TurtleBot2 Software}

Since the robot supports ROS(Robot Operating System) to communicate and execute instructions, the primary  mode of information exchange is the Subscriber/Publisher technique where the controller node reads subscribes to topics which have information about the surroundings and publishes data for other nodes as per requirement. The Twist message from 'geometry\_msgs' message type publishes messages to  '/cmd\_vel\_mux/input/navi' topic to control the movement of the robot's Kobuki base. 

Table 2 shows the messages used to maneuver the robot around. 

\begin{table}[ht]
\caption{Communication Messages}
\label{table_communications}
\begin{center}
\begin{tabular}{|l|l|l|}
\hline
\textbf{Control} & Topic & \textbf{Message} \\
\hline
Linear Velocity X & $ /cmd\_vel\_mux/input/navi $ & $linear.x$ \\
\hline
Linear Velocity Y & $/cmd\_vel\_mux/input/navi $ & $linear.y $\\
\hline
Angular Velocity Z & $/cmd\_vel\_mux/input/navi $& $angular.x $ \\
\hline
\end{tabular}
\end{center}
\end{table}

The controller node subscribes to several topics, namely, 'camera/depth/image\_raw', '/scan' and
'/camera/depth/points' providing the continuous depth cloud data points, horizontal laser scan data(array with distances of obstacle in the range of view) and complete point cloud visualization information respectively. It also fetches knowledge about the robot's current position in the world and its previous motion from related topics like 'joint\_states' and 'gazebo/link\_states'.

\begin{figure}[ht!]
  \centering
  \includegraphics[width = 8cm]{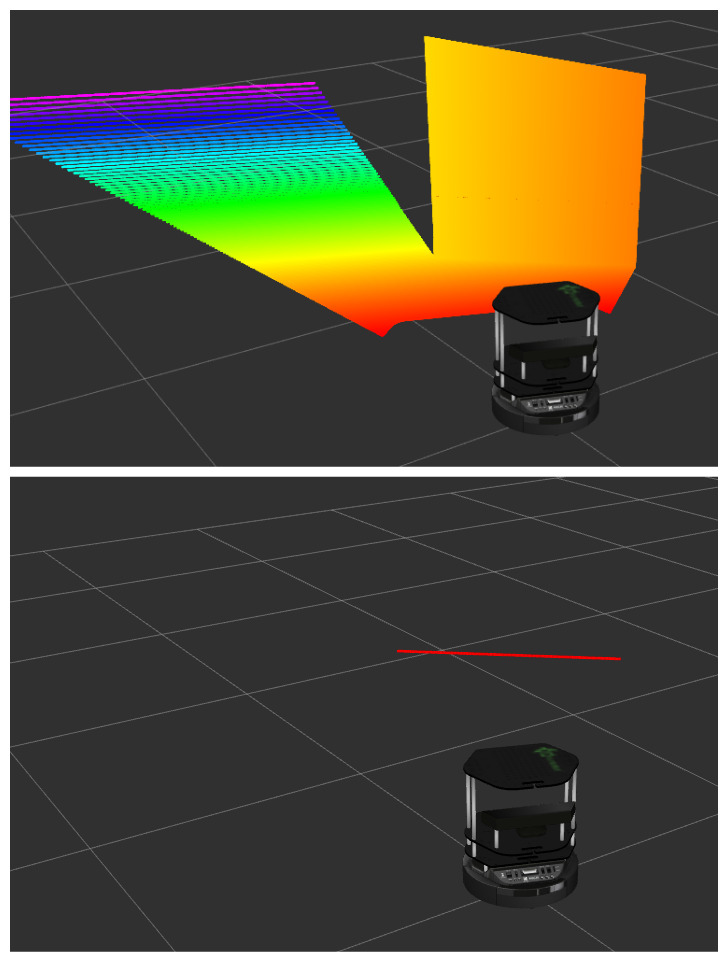}
   \caption{Data from Depth Cloud and LaserScan}
   \label{fig3}
\end{figure}

Figure \ref{fig3} shows the information obtained from Depth Cloud and Laser Scan data obtained from the sensor.   

The '/joint\_states' topic provides total distance traveled by each of the wheels(based on the revolutions) and the velocities of each individual wheel. The fuzzy logic controller is essentially supposed to determine the Cartesian velocities for the robot and feed the corresponding angular velocities to the wheels. However, since the robot already supports taking commands in Cartesian coordinates, the controller does not need to make those conversions. Finally, the position and orientation is published to the '/gazebo/link\_states' topic which has messages like Twist, Pose and reference frame information.

\subsection{Environmental Setup}
\begin{figure}[ht]
  \centering
  \includegraphics[width = 8cm]{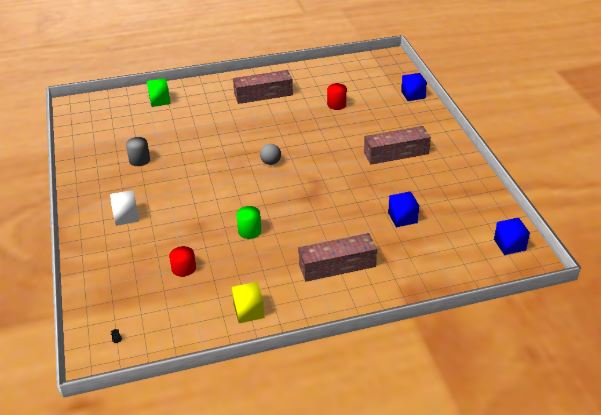}
   \caption{Customized Environment in Gazebo}
   \label{fig4}
\end{figure}

In order to test our controller performance we defined a customized environment in the Gazebo simulator as shown in Figure \ref{fig4}. The environment consists of objects of different shapes and sizes in order to increase the complexity of the data input from the Xion sensor. The objects are placed randomly. We can spawn our robot at any point in the environment and provide it with different goal positions to check the robustness of our controller. 

\subsection{Fuzzification of Kinect/Xion data}

\begin{figure}[ht]
  \centering
  \includegraphics[width = 8cm]{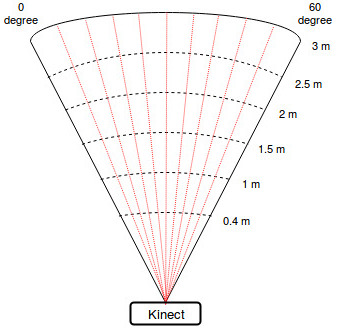}
   \caption{Fuzzification of Point Cloud data}
   \label{fig5}
\end{figure}

Figure \ref{fig5} shows the fuzzification process for data coming from the  Kinect/Xion sensor available on the TurtleBot2. Data is discretized based on the angular subsections of the depth image scan at every $3^{\circ}$. Thus, the total depth image is discretized in 20 subsections. Further, the depth values are discretized in subsections of around 0.5m . The depth data is thus divided in 5 sections ranging from 0.4m to 3m.\\
This discretized data is used to decide on the linear velocity values for left and right wheels which correspond to linear and angular velocity for the TurtleBot2 in our case.

\subsection{Implementation of the Fuzzy Logic}

\begin{figure}[ht!]
  \centering
  \includegraphics[width = 8cm]{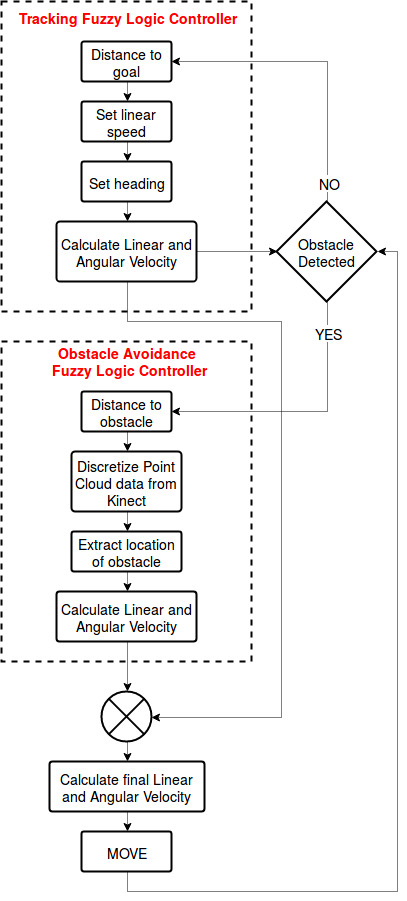}
   \caption{ FLC Implementation Flow Chart}
   \label{fig6}
\end{figure}

As shown in Figure \ref{fig6}, the implementation of our controller involves simultaneously running Tracking FLC and Obstacle Avoidance FLC. Each of them contribute towards the final decision on the linear velocities and the current direction of heading.

While the TFLC tries to adjust the robot's heading in the direction of the  target and set a linear speed that makes the robot move towards the goal, the OAFLC runs If-Else inference conditions on the obstacles encountered, depending upon their distance and angular position in robot's field of view. 
The OAFLC adjusts the heading and the linear speed together to just be able to dodge the obstacle with minimal effect in the previous speed/heading. This iterative process terminates after the robot reaches the target position. 

The final control signals could be obtained from the mathematical equation as:
\begin{center}
$(\dot{x}, \dot{\omega}_z) = x*(\dot{x}, \dot{\omega}_z)_{TFLC} + (1-x)*(\dot{x}, \dot{\omega}_z)_{OAFLC}$ \\
\end{center}

where, $\dot{x}$ and $\dot{\omega}_z$ represent the linear velocity in X direction and the angular velocity about the Z axis, for the robot. The equation provides the final commands to be sent to the robot which is a weighted sum of the same generated independently by the TFLC and the OAFLC. \\

\subsubsection{Obstacle Avoidance FLC}

The Obstacle avoidance Fuzzy Logic Controller works based on the sensor data for distance from obstacles. In case of TurtleBot 2, we use the Depth camera of Kinect to extract depth values at certain angles. The incoming depth values are also discretized as shown in Figure \ref{fig5} with several depths and angle discretizations forming an angular grid.

\begin{figure}[ht!]
  \centering
  \includegraphics[width = 8cm]{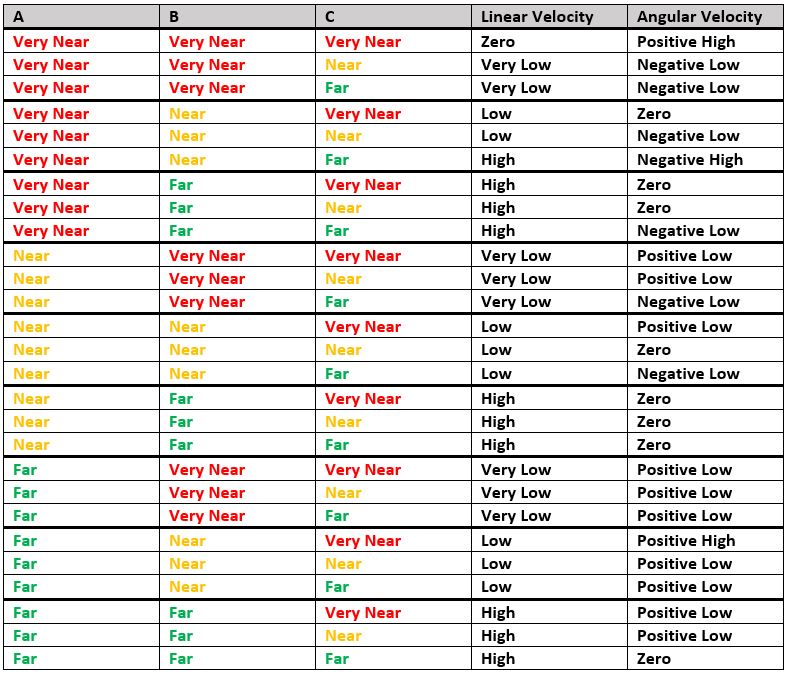}
   \caption{Rules for OAFLC}
   \label{fig9}
\end{figure}

The discretization in depth can also be changed based on the application and complexity of problem. This however changes the number of If-Else rules that are created based on the conditions imposed on each depth reading.

The proposed implementation here divides the angular range in 3 sections of 20 degrees (as the Kinect sensor has a range of $60^o$) each and depth is also discretized into 3 sections named as Very Near, Near, Far. Final inference rules change based on the combination of the three depth values and sections in which each of them falls. The If-Else conditions thus obtained are shown in Figure \ref{fig9} where columns A, B and C are represent the distance between the robot and the obstacle in those angular sections. 

\subsubsection{Tracking FLC}

Given that the TFLC takes the distance between the robot and target and the angular deviation between the robot's current heading and the line joining the robot to the target, the TFLC does not need any extra sensing setup. Proprioception from odometry data gives the current angular heading of the robot and the distance between the robot and target. 

\begin{figure}[ht!]
  \centering
  \includegraphics[width = 8cm]{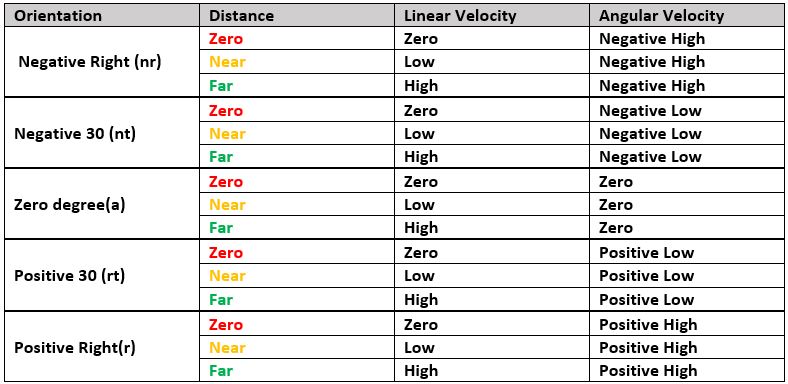}
   \caption{Rules for TFLC}
   \label{fig10}
\end{figure}

Angular heading deviation was fuzzified into 5 sections named Negative Right ($-90^o$), Negative Thirty ($-30^o$), Aligned ( $0^o$), Positive Thirty ($30^o$) and Positive Right ($90^o$) while the distance was fuzzified into simpler Zero, Near and Far sections. The resultant If-Else rules generated are shown in Figure \ref{fig10}

\begin{figure}[ht!]
  \centering
  \includegraphics[width = 8cm]{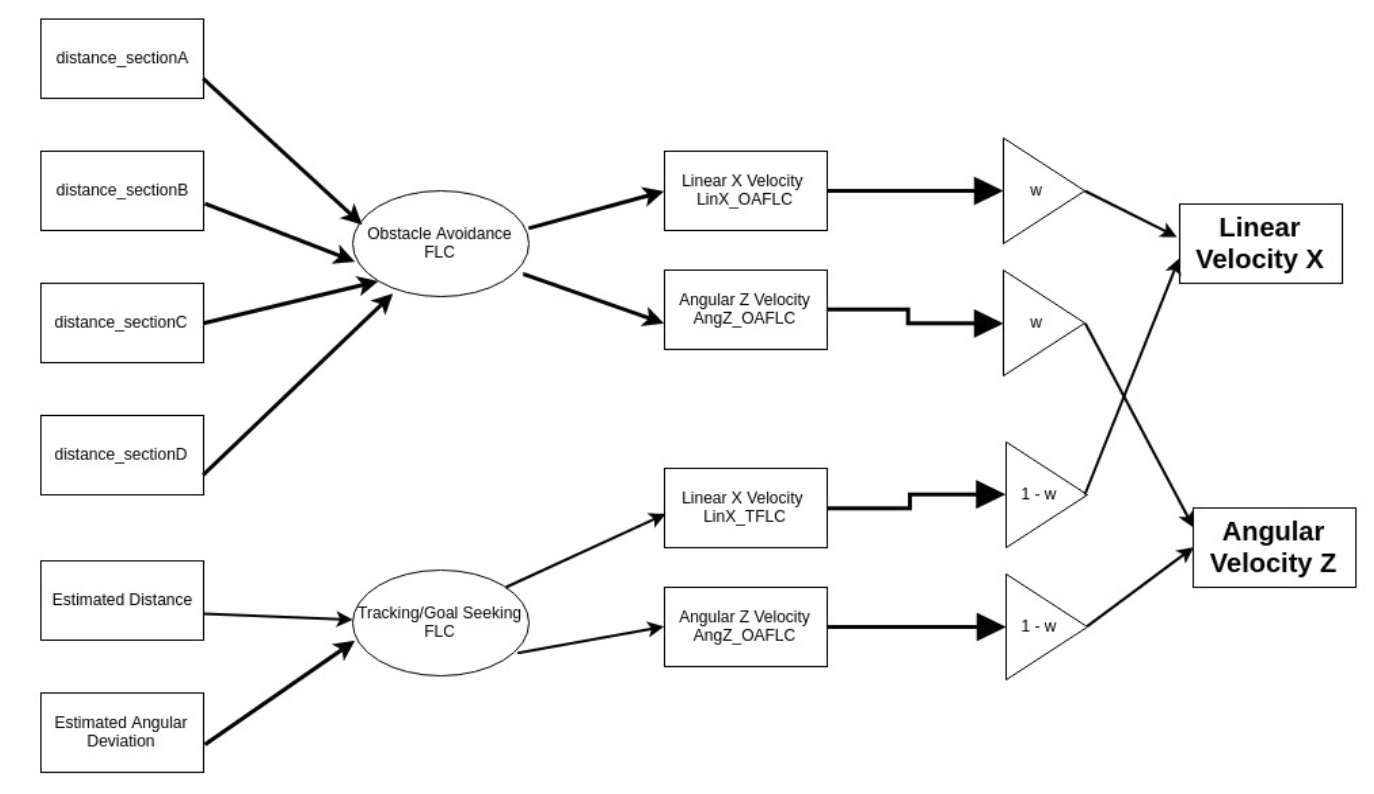}
   \caption{Behavior Fusion and Flow of Control}
   \label{fig12}
\end{figure}

The final behavior fusion from the two FLCs is shown in Figure \ref{fig12} which shows final weighted sum of the same as the final commands sent to the robot for its motion. 

\section{Experimental Results}

\begin{figure}[ht!]
  \centering
  \includegraphics[width = 8cm]{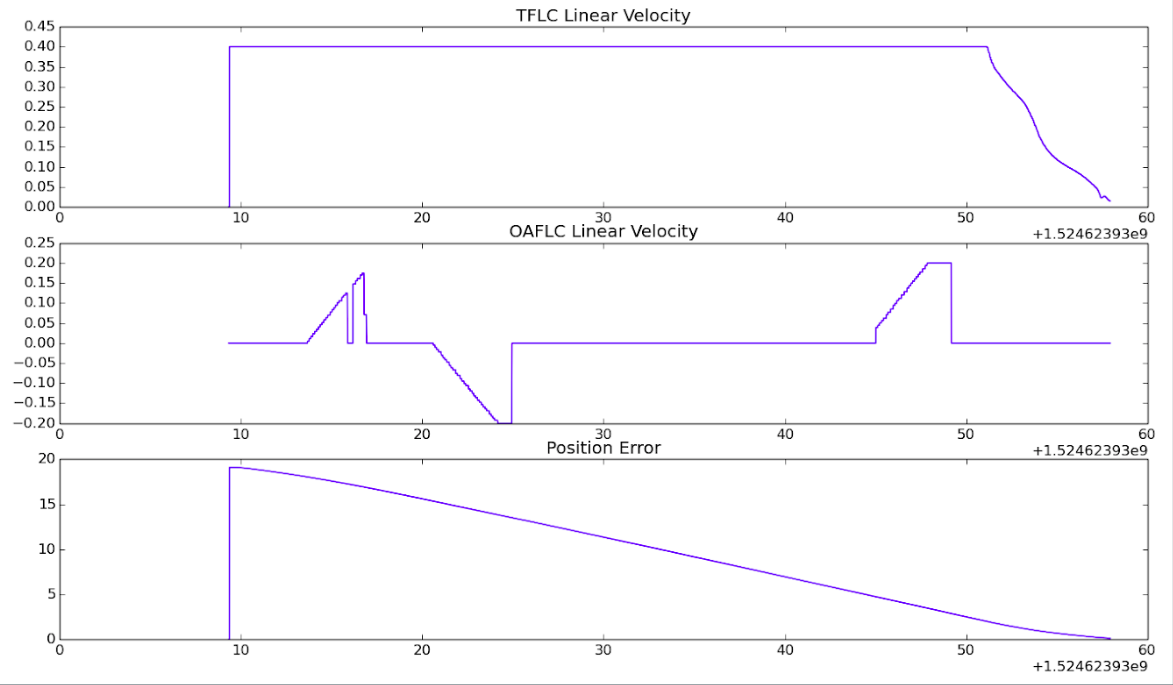}
   \caption{Predicted Linear Velocities by the TFLC and OAFLC and convergence of position error over time}
   \label{fig11}
\end{figure}

The proposed FLC methodology delivered satisfactory results during implementation in the simulation environment. Figure \ref{fig11} shows variations in linear velocities while the robot tries to traverse from a far off start point to reach the target. 

As evident in the results, over time, the TFLC predicts highest linear velocity when the robot is far off and then gradually decreases as it nears the target. The peaks in OAFLC predicted linear velocity distribution suggests that it encountered obstacles at those time-steps and thus predicted a changed linear velocity at a changed angular orientation to dodge the obstacle. 

The devised controller was tested for varying complexity in simulation environment as well as different start and target positions. TurtleBot2 was able to reach those locations within very acceptable time period. The controller performance was satisfactory for all the testing situations. Two of the results have been recorded and the video showing the same can be found here at \url{https://www.youtube.com/watch?v=GUEN4Orpb2A} and \url{https://www.youtube.com/watch?v=4fj4q-swg0U}

Since the proposed technique does not make the robot track any pre-defined trajectory or constraints, the results do not have any comparative graphical content but rather the above videos show complete implementations. 

\section{Future Work}

We propose using techniques like Genetic algorithm and Particle-Swarm optimization to improve the performance of our system. Genetic algorithm is an evolutionary algorithm that uses biological operators like mutation, crossovers, elitism and culling. The algorithm tunes the fuzzy control rules and tries to make the system resemble an ideal control system. The tuning method fits the fuzzy rules' membership functions with the FIS (Fuzzy Inference System) and the defuzzification process. In the end, the method extracts best membership functions for the process. Particle-Swarm optimization technique is another similar iterative evolutionary algorithm that improves a candidate solution by making it "fly" through the problem space following the current best solution.

\section{Conclusion}

We were able to get satisfactory performance for robot navigation in unknown environments with no prior knowledge about obstacles. The proposed FLC implementation 
using native localization from the simulation environment which could be eliminated easily. We also faced the following problems in the project.

\subsection{ Problems Faced}
As we are using the Kinect sensor on the TurtleBot2 we are facing the following range limitations:\\
\begin{enumerate}
	\item The Xion gives an angular range of 58.5 degrees divided with a central axis. Therefore it limits the visibility range and cannot detect obstacles out of that angular range. Therefore, unlike the LIDAR on TurtleBot3, TurtleBot2 lacks a 360 degree view and the controller shall have limited sensing which might affect the performance we fear. 
    \item We get the depth image from the Xion which gives data from range 40cm to 3 meters in the near mode. Hence, any obstacle between this range can be detected. But this creates a limitation for detection of obstacles nearby the robot, at a distance less than 40cm which becomes a blind spot and makes the controller limited performance on sudden appearance of obstacles, difficult. 
\end{enumerate}

\bibliographystyle{plain}
\bibliography{references}

\end{document}